\documentclass[11pt,a4paper]{article}

\usepackage[hyperref]{acl2020}
\usepackage{times}
\usepackage{latexsym}

\usepackage[utf8]{inputenc}
\usepackage{enumitem}
\usepackage{url}
\usepackage{booktabs}
\usepackage{graphicx}

\usepackage{amsmath}
\usepackage{amssymb}
\usepackage{todonotes}
\usepackage{mathtools}
\usepackage{siunitx}
\usepackage{multirow}
\usepackage{multicol}

\usepackage{caption}
\usepackage{subcaption}

\usepackage{microtype}

\aclfinalcopy 
\def\aclpaperid{1514} 

\title{MultiQT: Multimodal Learning \\for Real-Time Question Tracking in Speech}

\author{
    Jakob D. Havtorn \quad Jan Latko \quad Joakim Edin \quad Lasse Borgholt  \quad Lars Maaløe \\ {\bf Lorenzo Belgrano \quad Nicolai F. Jacobsen \quad Regitze Sdun \quad Željko Agić}\\
    Corti\\
    Store Strandstræde 21, 4\\
    1255 Copenhagen K, Denmark\\
    {\tt jdh@corti.ai}
    }

\date{}

\begin{document}
\maketitle
\begin{abstract}
We address a challenging and practical task of labeling questions in speech in real time during telephone calls to emergency medical services in English, which embeds within a broader decision support system for emergency call-takers. We propose a novel multimodal approach to real-time sequence labeling in speech. Our model treats speech and its own textual representation as two separate modalities or views, as it jointly learns from streamed audio and its noisy transcription into text via automatic speech recognition. Our results show significant gains of jointly learning from the two modalities when compared to text or audio only, under adverse noise and limited volume of training data. The results generalize to medical symptoms detection where we observe a similar pattern of improvements with multimodal learning.
\end{abstract}

\section{Introduction}

Our paper addresses the challenge of learning to discover and label questions in telephone calls to emergency medical services in English. The task is demanding in two key aspects:
\begin{enumerate}[noitemsep,wide=0pt]
    \item {\bf Noise:} A typical phone call to an emergency medical service differs significantly from data within most standard speech datasets. Most importantly, emergency calls are noisy by nature due to very stressful conversations conveyed over poor telephone lines. Automatic speech recognition (ASR) and subsequent text processing quickly becomes prohibitive in such noisy environments, where word error rates (WER) are significantly higher than for standard benchmark data~\cite{han2017deep}. For this reason, we propose a sequence labeler that makes use of two modalities of a phone call: audio and its transcription into text by utilizing an ASR model. Hereby we create a multimodal architecture that is more robust to the adverse conditions of an emergency call.
    \item {\bf Real-time processing:} Our model is required to work incrementally to discover questions in real time within incoming streams of audio in order to work as a live decision support system. At runtime, no segmentation into sub-call utterances such as phrases or sentences is easily available. The lack of segmentation coupled with the real-time processing constraint makes it computationally prohibitive to discover alignments between speech and its automatic transcription. For these reasons, we cannot utilize standard approaches to multimodal learning which typically rely on near-perfect cross-modal alignments between short and well-defined segments~\cite{baltruvsaitis2018multimodal}.
\end{enumerate}

\begin{figure}[t]
    \centering
    \includegraphics[width=\columnwidth]{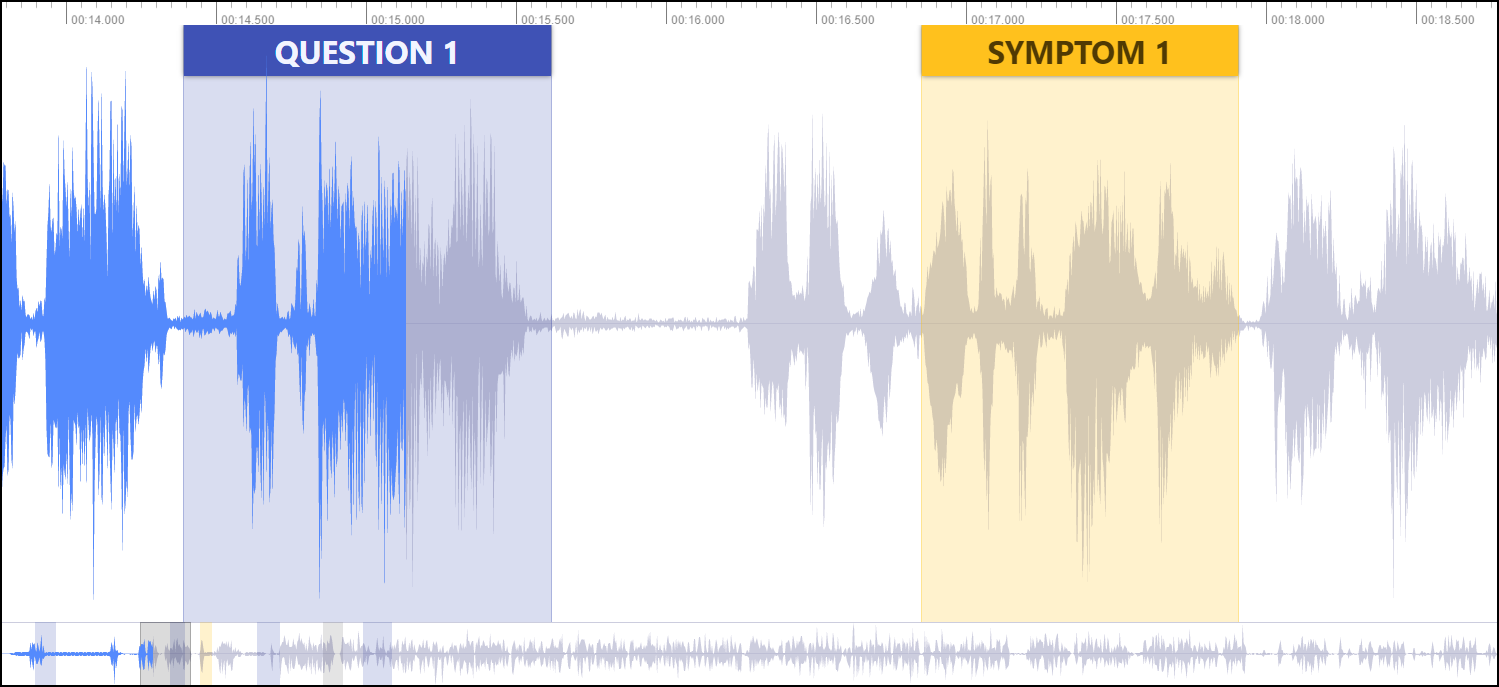}
    \caption{A speech sequence from our phone call dataset. Two audio segments are highlighted: a question (in blue) and a reported symptom (in yellow).}
    \label{fig:screenshot}
\end{figure}

\paragraph{Context and relevance.} Learning to label sequences of text is one of the more thoroughly explored topics in natural language processing. In recent times, neural networks are applied not only to sequential labeling like part-of-speech tagging~\cite{plank-etal-2016-multilingual} or named entity recognition~\cite{ma-hovy-2016-end}, but also to cast into a labeling framework otherwise non-sequential tasks such as syntactic parsing~\cite{gomez-rodriguez-vilares-2018-constituent,strzyz-etal-2019-viable}.

By contrast, assigning labels to audio sequences of human speech is comparatively less charted out. When addressed, speech labeling typically adopts a solution by proxy, which is to automatically transcribe speech into text, and then apply a text-only model~\cite{surdeanu2005named,molla-etal-2007-named,eidelman-etal-2010-lessons}. The challenge then becomes not to natively label speech, but to adapt the model to adverse conditions of speech recognition error rates. Such models typically feature in end-to-end applications such as dialogue state tracking~\cite{henderson2014second,ram2018alexa}. Recent advances in end-to-end neural network learning offer promise to directly label linguistic categories from speech alone~\cite{ghannay2018end}. From another viewpoint, multimodal learning is successfully applied to multimedia processing where the modalities such as text, speech, and video are closely aligned. However, contributions there typically feature classification tasks such as sentiment analysis and not finer-grained multimedia sequence labeling~\cite{zadeh-etal-2017-tensor}.

\paragraph{Our contributions.} We propose a novel neural architecture to incrementally label questions in speech by learning from its two modalities or views: the native audio signal itself and its transcription into noisy text via ASR.

\begin{enumerate}[noitemsep,leftmargin=*]
    \item Our model utilizes the online temporal alignment between the input audio signal and its raw ASR transcription. By taking advantage of this fortuitous real-time coupling, we avoid having to learn the multimodal alignment over the entire phone call and its transcript, which would violate the real-time processing constraint that is crucial for decision support.
    \item We achieve consistent and significant improvements from learning jointly from the two modalities compared to ASR transcriptions and audio only. The improvements hold across two inherently different audio sequence labeling tasks.
    \item Our evaluation framework features a challenging real-world task with noisy inputs and real-time processing requirements. Under this adversity, we find questions and medical symptoms in emergency phone calls with high accuracy. Our task is illustrated in~\autoref{fig:screenshot}.
\end{enumerate}

\section{Multimodal speech labeling}

\begin{figure}[t]
    \centering
    \includegraphics[width=\columnwidth]{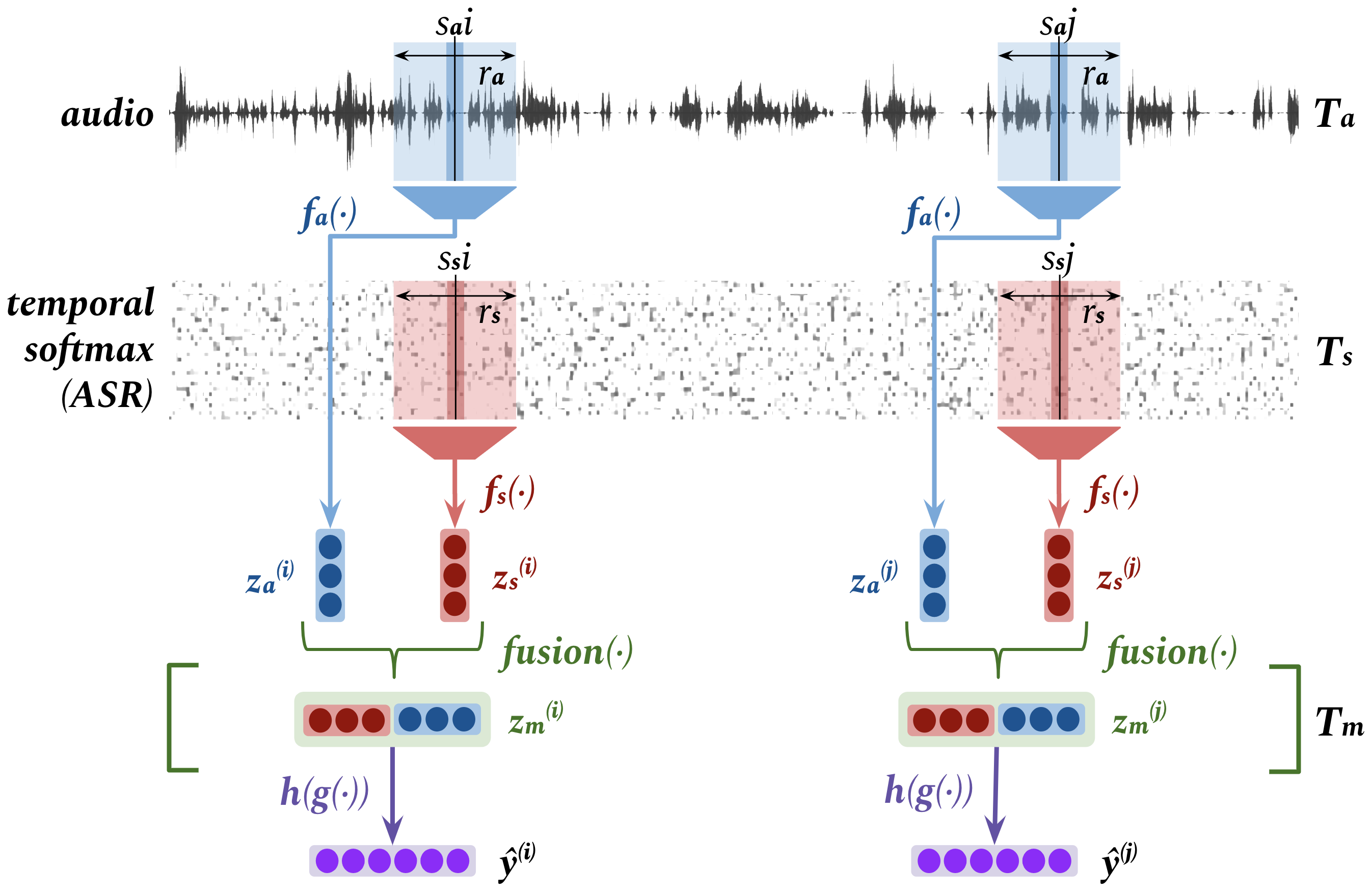}
    \caption{MultiQT model illustration for two timesteps $i$ and $j$. We depict the convolutional transformations $f_a$ and $f_s$ of the audio and character temporal softmax inputs into the respective modality encodings $\mathbf{z}_a^{(i)}$ and $\mathbf{z}_s^{(i)}$, along with the corresponding receptive fields and strides: $r_a, s_a$ and $r_s, s_s$. The convolutions are followed by multimodal $\mathrm{fusion}$ and finally dense layers $g$ and $h$ to predict the question labels $\mathbf{\hat{y}}^{(i)}$ and $\mathbf{\hat{y}}^{(j)}$.}
    \label{fig:fusion-arch}
\end{figure}

We define the multimodal speech labeler MultiQT as a combination of three neural networks that we apply to a number of temporal input modalities.
In our case, we consider speech and associated machine transcripts as the separate modalities or views. The model is illustrated in \autoref{fig:fusion-arch}.

To obtain temporal alignment between speech and text, we propose a simple approach that uses the output of an ASR system as the textual representation. Here, we take the ASR to be a neural network trained with the connectionist temporal classification (CTC) loss function~\citep{graves-etal-2006-connectionist}. Given audio, it produces a temporal softmax of length $T_s$ with a feature dimension defined as a categorical distribution, typically over characters, words or subword units, per timestep.

We refer to a sequence of input representations of the audio modality as $(\mathbf{x}_a^{(t)})_{t\in[1..T_a]}$ and of the textual modality as $(\mathbf{x}_s^{(t)})_{t\in[1..T_s]}$. From the input sequences we compute independent unimodal representations denoted by $\mathbf{z}_a^{(t)}$ and $\mathbf{z}_s^{(t)}$ by applying two unimodal transformations denoted by $f_a$ and $f_s$, respectively. Each of these transformations is parameterized by a convolutional neural network with overall temporal strides $s_a$ and $s_s$ and receptive fields $r_a$ and $r_s$. With $T_m$ as length of the resulting unimodal representations:
\begin{equation}
    \begin{split}
        \mathbf{z}_a^{(t)} &= f_a\left( \left(\mathbf{x}_a^{(i)}\right)_{i=s_a t-r_{a,l}}^{s_a t+r_{a,r}} \right)\\
        \mathbf{z}_s^{(t)} &= f_s\left( \left(\mathbf{x}_s^{(i)}\right)_{i=s_s t-r_{s,l}}^{s_s t+r_{s,r}} \right),
    \end{split}
\end{equation}
for $t\in[1..T_m]$, where $r_{a,l}$, $r_{a,r}$, $r_{s,l}$ and $r_{s,r}$ are the left and right half receptive fields of $f_a$ and $f_s$, respectively. For $f_a$, $r_{a,l}=\left\lfloor (r_a - 1)/2 \right\rfloor$ and $r_{a,r}=\left\lceil (r_a - 1)/2 \right\rceil$ and similarly for $f_s$. For $i<1$ and $i>T_a$ we define $\mathbf{x}_a^{(i)}$ and $\mathbf{x}_s^{(i)}$ by zero padding, effectively padding with half the receptive field on the left and right sides of the input. This then implies that $T_m = \left\lfloor T_a / s_a \right\rfloor = \left\lfloor T_s / s_s \right\rfloor$ which constrains the strides according to $T_a$ and $T_s$ and functions as ``same padding''. This lets us do convolutions without padding the internal representations for each layer in the neural networks, which in turn allows for online streaming. 

To form a joint multimodal representation from $\mathbf{z}_a^{(t)}$ and $\mathbf{z}_s^{(t)}$ we join the representations along the feature dimension. In the multimodal learning litterature such an operation is sometimes called fusion~\citep{zadeh-etal-2017-tensor}. We denote the combined multimodal representation by $\mathbf{z}_m^{(t)}$ and obtain it in a time-binded manner such that for a certain timestep $\mathbf{z}_m^{(t)}$ only depends on $\mathbf{z}_a^{(t)}$ and $\mathbf{z}_s^{(t)}$,
\begin{equation}
    \mathbf{z}_m^{(t)} = \text{fusion}\left(\mathbf{z}_a^{(t)}, \mathbf{z}_s^{(t)}\right).
\end{equation}
In our experiments $\text{fusion}(\cdot)$ either denotes a simple concatenation, $[\mathbf{z}_a^{(t)};\mathbf{z}_s^{(t)}]$, or a flattened outer product, $[1\;\mathbf{z}_a^{(t)}] \otimes [1\; \mathbf{z}_s^{(t)}]$. The latter is similar to the fusion introduced by \citet{zadeh-etal-2017-tensor}, but we do not collapse the time dimension since our model predicts sequential labels. 

Finally, $\mathbf{z}_m^{(t)}$ is transformed before projection into the output space:
\begin{align}
    \mathbf{z}_y^{(t)} &= g\left(\mathbf{z}_m^{(t)}\right), \label{eq: Output latent representation transformation} \\
    \hat{\mathbf{y}}^{(t)} &= h\left(\mathbf{z}_y^{(t)}\right), \label{eq: Output transformation}
\end{align}
where $g$ is a fully connected neural network and $h$ is a single dense layer followed by a softmax activation such that $\hat{\mathbf{y}}^{(t)}\in\mathbb{R}^{K}$ is a vector of probabilities summing to one for each of the $K$ output categories. The predicted class is $\arg\max(\hat{\mathbf{y}}^{(t)})$.

\subsection{Objective functions}

In general, the loss is defined as a function of all learnable parameters $\Theta$ and is computed as the average loss on $M$ examples in a mini-batch. 
We denote by $\{\mathcal{X}_a, \mathcal{X}_s\}$ a dataset consisting of $N$ pairs of input sequences of each of the two modalities.
As short-hand notation, let $\mathbf{X}_a^{(n)}$ refer to the $n$'th audio sequence example
in $\mathcal{X}_a$ and similarly for $\mathbf{X}_s^{(n)}$. The mini-batch loss is then
\begin{equation}\label{eq: Objective function}
    \begin{aligned}
        & \mathcal{L}\left( \Theta; \left\{ \mathbf{X}_a^{(n)}, \mathbf{X}_s^{(n)} \right\}_{n\in\mathcal{B}_i} \right) = \\
        & \quad\quad \frac{1}{M}\sum_{n\in\mathcal{B}_i} \mathcal{L}^{(n)}\left( \Theta; \mathbf{X}_a^{(n)}, \mathbf{X}_s^{(n)} \right),
    \end{aligned}
\end{equation}
where $\mathcal{B}_i$ is an index set uniformly sampled from $[1..N]$ which defines the $i$'th batch of size $|\mathcal{B}_i|=M$.

The loss for each example, $\mathcal{L}^{(n)}$, is computed as the time-average of the loss per timestep,
\begin{equation}
    \begin{aligned}
        & \mathcal{L}^{(n)} \left( \Theta; \mathbf{X}_a^{(n)}, \mathbf{X}_s^{(n)} \right) = \\
        & \quad\quad \frac{1}{T} \sum_{t=1}^T \mathcal{L}^{(n,t)} \left(  
            \Theta;
            \mathbf{X}_a^{(n,\mathbf{t}_a)},
            \mathbf{X}_s^{(n,\mathbf{t}_s)}
        \right),
    \end{aligned}
\end{equation}
where $\mathbf{t}_a=[s_at-r_{a,l}\;..\;s_at+r_{a,r}]$ and similarly for $\mathbf{t}_s$ since the dependency of the loss per timestep is only on a limited timespan of the input.
The loss per timestep is defined as the categorical cross-entropy loss between the softmax prediction $\hat{\mathbf{y}}^{(t)}$ and the one-hot encoded ground truth target $\mathbf{y}^{(t)}$,
\begin{align*}
    \mathcal{L}^{(n, t)} \left( \Theta; \mathbf{X}_a^{(n,\mathbf{t}_a)}, \mathbf{X}_s^{(n,\mathbf{t}_s)} \right) = \sum^{K}_{k=1}y_k^{(t)} \log(\hat{y}_k^{(t)}).
\end{align*}
The full set of learnable parameters $\Theta$ is jointly optimized by mini-batch stochastic gradient descent.

\begin{table*}[t]
    \centering
    \resizebox{\textwidth}{!}{  
        \begin{tabular}{cllcc}
        \toprule
        Label & Description & Example & Count & Fraction\\
        \midrule
        Q1 & Question about the address of the incident. & \textit{What’s the address?} & 663 & 26.3\%\\
        Q2 & Initial question of the call-taker to begin assessing the situation. & \textit{What’s the problem?} & 546 & 21.6\%\\
        Q3 & Question about the age of the patient. & \textit{How old is she?} & 537 & 21.3\%\\
        Q4 & All questions related to patient's quality of breathing. & \textit{Is she breathing in a normal pattern?} & 293 & 11.6\%\\
        Q5 & All question about patient's consciousness or responsiveness. & \textit{Is he conscious and awake?} & 484 & 19.2\%\\
        \bottomrule
        \end{tabular}
    }
    \caption{Explanation and prevalence of the questions used for the experiments.}
    \label{tab:question_classes}
\end{table*}

\subsection{Multitask objective}

In addition to the loss functions defined above, we also consider multitask training. This has been reported to improve performance in many different domains by including a suitably related auxiliary task~\cite{bingel-sogaard-2017-identifying,martinez-alonso-plank-2017-multitask}.

For the task of labelling segments in the input sequences as pertaining to annotations from among a set of $K-1$ positive classes and one negative class, we propose the auxiliary task of binary labelling of segments as pertaining to either the negative class or any of the $K-1$ positive classes. For question tracking, this amounts to doing binary labelling of segments that are questions of any kind. The hope is that this will make the training signal stronger since the sparsity of each of the classes, e.g. questions, is reduced by collapsing them into one shared class.
 
We use the same loss function as above, but with the number of classes reduced to $K=2$. The total multitask loss is a weighted sum of the $K$-class loss and the binary loss:
\begin{equation}
    \mathcal{L}_{\text{MT}}^{(n, t)} = 
    \beta \mathcal{L}_{\text{binary}}^{(n, t)} + 
    (1 - \beta) \mathcal{L}_{}^{(n, t)}.
\end{equation}
The tunable hyperparameter $\beta\in[0,1]$ interpolates the task between regular $K$-class labeling for $\beta=0$ and binary classification for $\beta=1$.

\section{Data}

Our dataset consists of 525 phone calls to an English-speaking medical emergency service. The call audio is mono-channel, PCM-encoded and sampled at \SI{8000}{Hz}.
The duration of the calls has the mean of \SI{166}{s} (st. dev. \SI{65}{s}, IQR \SI{52}{s}). All calls are manually annotated for questions by trained native English speakers. Each question is annotated with its start and stop time and assigned with one of 13 predefined question labels or an additional label for any question that falls outside of the 13 categories. \autoref{fig:screenshot} illustrates these annotations. We observe an initial inter-annotator agreement of $\alpha=0.8$~\cite{krippendorff2018content}. 
Each call has been additionally corrected at least once by a different annotator to improve the quality of the data. On average it took roughly 30 minutes to annotate a single call. For our experiments, we choose the five most frequent questions classes, which are explained in \autoref{tab:question_classes}. Out of 24 hours of calls, the questions alone account for only 30 minutes (roughly 2\%) of audio. For the experiments we use 5-fold cross-validation stratified by the number of questions in each call, such that calls of different lengths and contents are included in all folds.

We test our model on an additional speech sequence labeling challenge: tracking mentions of medical symptoms in incoming audio. By using another task we gauge the robustness of MultiQT as a general sequence labeling model and not only a question tracker, since symptom utterances in speech carry inherently different linguistic features than questions. As our question-tracking data was not manually labeled for symptoms, we created silver-standard training and test sets automatically by propagating a list of textual keywords from the ground truth human transcripts back onto the audio signal as time stamps with a rule-based algorithm. The initial list contained over 40 medical symptoms, but in the experiment we retain the most frequent five: state of consciousness, breathing, pain, trauma, and hemorrhage.

The utterances that we track are complex phrases with a high variance: There are many different ways to express a question or a medical symptom in conversation. This linguistic complexity sets our research apart from most work in speech labeling which is much closer to exact pattern matching~\cite{salamon2017deep}.

\section{Experiments}

\subsection{Setup}
\paragraph{Inputs.}
The audio modality is encoded using $40$ log-mel features computed with a window of \SI{0.02}{s} and stride \SI{0.01}{s}.

The textual modality is formed by application of an ASR system to the audio modality. In all reported experiments, only ASR outputs are used and never human transcriptions, both in training and evaluation. 
The audio input to the ASR is encoded in the same way as described above. The ASR available to us has a purely convolutional architecture similar to the one in~\cite{Collobert2016} with an overall stride of $2$. For MultiQT, this means that $T_a=2T_s$.
The ASR is trained on 600 hours of phone calls to medical emergency services in English from the same emergency service provider as the question and symptoms tracking datasets. Both of these are contained in the ASR test set.
The ASR is trained using the connectionist temporal classification (CTC) loss function~\citep{graves-etal-2006-connectionist} and has a character error rate of \SI{~14}{\%} and a word error rate of \SI{~31}{\%}.
Its feature dimension is 29 which corresponds to the English alphabet including apostrophe, space and a blank token for the CTC loss.

\paragraph{Systems.} The basic version of MultiQT uses a single softmax cross-entropy loss function and forms a time-bound multimodal representation by concatenating the unimodal representations. We then augment this model in three ways:
\begin{enumerate}[noitemsep,leftmargin=*]
    \item MultiQT-TF: tensor fusion instead of concatenation following~\citet{zadeh-etal-2017-tensor},
    \item MultiQT-MT: auxiliary binary classification with $\beta=0.5$, 
    \item MultiQT-TF-MT: combination of 1 and 2.
\end{enumerate}

\paragraph{Baselines.}
MultiQT can easily be adapted to a single modality by excluding the respective convolutional transformation $f_a$ or $f_s$. 
For example, MultiQT can be trained unimodally on audio by removing $f_s$ and then defining $\mathbf{z}_m^{(t)} = \mathbf{z}_a^{(t)}$ instead of concatenation or tensor fusion.
We baseline the multimodal MultiQT models against versions trained unimodally on audio and text. We also compare MultiQT to two distinct baseline models:
\begin{enumerate}[noitemsep,leftmargin=*]
    \item Random forest (RF)
    \item Fully connected neural network (FNN)
\end{enumerate} 
Contrary to MultiQT, the baselines are trained to classify an input sequence into a single categorical distribution over the labels. At training, the models are presented with short segments of call transcripts in which all timesteps share the same label such that a single prediction can be made. The baselines are trained exclusively on text and both models represent the windowed transcript as a TF-IDF-normalized bag of words similar to~\citet{zhang2015character}. The bag of words uses word uni- and bigrams, and character tri-, four- and five-grams with 500 of each selected by $\chi^2$-scoring between labels and transcripts on the training set.

\paragraph{Hyperparameters.}
We use 1D convolutions for $f_a$ and $f_s$. For $f_a$ we use three layers with kernel sizes of 10, 20 and 40, filters of 64, 128 and 128 units and strides of 2, 2 and 2 in the first, second and third layer, respectively. For $f_s$ we use two layers with kernel sizes of 20 and 40, filters of 128 and 128 units and strides of 2 and 2. Before each nonlinear transformation in both $f_a$ and $f_s$ we use batch normalization~\citep{ioffe-etal-2015-batch} with momentum $0.99$ and trainable scale and bias, and we apply dropout~\citep{srivastava-2014-dropout} with a dropout rate of 0.2. For $g$ we use three fully connected layers of 256 units each and before each nonlinear transformation we use batch normalization as above and apply dropout with a dropout rate of 0.4.
We $l_2$ regularize all learnable parameters with a weighting of $0.1$.

The FNN model uses the same classifier as is used for $g$ in MultiQT with a dropout rate of 0.3 and an $l_2$ regularization factor of $0.05$. 

All neural models are trained with the Adam optimizer~\citep{kingma-etal-2015-adam} using a learning rate of $1\times10^{-4}$, $\beta_1=0.9$ and $\beta_2=0.999$ and batch size 6 except for those with tensor fusion which use a batch size of 1 due to memory constraints. Larger batch sizes were prohibitive since we use entire calls as single examples but results were generally consistent across different batch sizes. All hyperparameters were tuned manually and heuristically. It takes approximately one hour to train the base MultiQT model on one NVIDIA GeForce GTX 1080 Ti GPU card.

\begin{table*}[t]
    \centering
    \resizebox{0.9\textwidth}{!}{
        \begin{tabular}{llcccccc}
        \toprule
        & & \multicolumn{3}{c}{{\sc instance}} & \multicolumn{3}{c}{{\sc timestep}}\\ 
        \cmidrule(l{2pt}r{2pt}){3-5} 
        \cmidrule(l{2pt}r{2pt}){6-8}
        Model & Modality & P & R & F1 & P & R & F1\\
        \cmidrule(l{2pt}r{2pt}){1-2} 
        \cmidrule(l{2pt}r{2pt}){3-5} 
        \cmidrule(l{2pt}r{2pt}){6-8}
        RF-BOW & T & 61.8$\pm$3.5 & 88.5$\pm$0.9 & 72.2$\pm$2.2 & 39.3$\pm$1.1 & 70.4$\pm$1.0 & 48.1$\pm$1.0 \\ 
        FNN-BOW & T & 42.2$\pm$1.4 & 92.8$\pm$0.6 & 57.5$\pm$1.3 & 38.1$\pm$0.7 & 71.0$\pm$1.7 & 46.9$\pm$0.8 \\ 
        \cmidrule(l{2pt}r{2pt}){1-2} 
        \cmidrule(l{2pt}r{2pt}){3-5} 
        \cmidrule(l{2pt}r{2pt}){6-8}
        MultiQT & A & 87.4$\pm$1.9 & 60.6$\pm$4.0 & 70.3$\pm$3.1 & 79.2$\pm$1.3 & 57.8$\pm$3.3 & 65.0$\pm$2.4 \\ 
        MultiQT & T & 84.2$\pm$1.6 & 78.5$\pm$2.8 & 81.1$\pm$2.0 & 78.8$\pm$1.2 & 69.4$\pm$2.0 & 73.5$\pm$1.3 \\ 
        MultiQT & A+T & 83.6$\pm$2.2 & 83.3$\pm$2.5 & \textbf{83.3$\pm$1.6} & 75.7$\pm$2.2 & 73.8$\pm$2.3 & \textbf{74.5$\pm$1.3} \\ 
        \cmidrule(l{2pt}r{2pt}){1-2} 
        \cmidrule(l{2pt}r{2pt}){3-5} 
        \cmidrule(l{2pt}r{2pt}){6-8}
        MultiQT-MT & A & 84.6$\pm$5.1 & 57.4$\pm$3.9 & 66.2$\pm$2.9 & 77.7$\pm$5.6 & 56.0$\pm$2.8 & 62.8$\pm$2.0 \\ 
        MultiQT-MT & T & 81.9$\pm$1.1 & 80.6$\pm$2.8 & 81.0$\pm$1.8 & 75.9$\pm$1.5 & 71.2$\pm$2.4 & 73.3$\pm$1.7 \\ 
        MultiQT-MT & A+T & 85.2$\pm$2.7 & 83.2$\pm$1.2 & \textbf{84.1$\pm$2.0} & 78.5$\pm$2.5 & 74.0$\pm$0.7 & \textbf{76.0$\pm$1.1} \\ 
        \cmidrule(l{2pt}r{2pt}){1-2} 
        \cmidrule(l{2pt}r{2pt}){3-5} 
        \cmidrule(l{2pt}r{2pt}){6-8}
        MultiQT-TF & A+T & 85.0$\pm$1.8 & 83.3$\pm$2.6 & \textbf{83.9$\pm$1.7} & 78.9$\pm$2.1 & 75.2$\pm$2.3 & \textbf{76.7$\pm$1.2} \\ 
        MultiQT-TF-MT & A+T & 85.1$\pm$3.2 & 83.1$\pm$1.6 & 83.8$\pm$1.7 & 78.7$\pm$3.7 & 75.0$\pm$1.6 & 76.5$\pm$1.4 \\ 
        \bottomrule
        \end{tabular}
    }
    \caption{Question tracking results on audio (A) and text (T) modalities with variations of MultiQT using modality concatenation (MultiQT) or tensor fusion (MultiQT-TF) and the auxiliary task (MultiQT-MT). The evaluation metrics are precision (P), recall (R), and (F1) at the macro level per {\sc timestep} or {\sc instance}. We report means and standard deviations for five-fold cross-validation runs. All F1 differences are statistically significant at $p<0.001$, save for between MulitQT [T] \& MulitQT-MT [T], and MulitQT [A+T] \& MulitQT-TF-MT [A+T] ($p\approx0.64$). We employ the approximate randomization test with $R=1000$ and Bonferonni correction~\cite{dror-etal-2018-hitchhikers}. Bold face indicates the highest F1 score within each metric and MultiQT model group.}
    \label{tab:results}
\end{table*}

\paragraph{Evaluation.} For each model we report two F1 scores with respective precisions and recalls macro-averaged over the classes.
\begin{itemize}[noitemsep,leftmargin=*]
    \item[--] {\sc timestep:} For each timestep, the model prediction is compared to the gold label. The metrics are computed per timestep and micro-averaged over the examples.
    This metric captures the model performance in finding and correctly classifying entire audio segments that represent questions and is sensitive to any misalignment.
    \item[--] {\sc instance:} A more forgiving metric which captures if sequences of the same label are found and correctly classified with acceptance of misalignment. Here, the prediction counts as correct if there are at least five consecutive correctly labeled time steps within the sequence, as a heuristic to avoid ambiguity between classes. This metric also excludes the non-question label.
\end{itemize}
The baseline models are evaluated per {\sc timestep} by labeling segments from the test set in a sliding window fashion. The size of the window varies from 3 to 9 seconds to encompass all possible lengths of a question with the stride set to one word. Defining the stride in terms of words is possible because the ASR produces timestamps for the resulting transcript per word.

\subsection{Results}

\paragraph{Labeling accuracy.} The results are presented in \autoref{tab:results}. They show that for any model variation, the best performance is achieved when using both audio and text. The model performs the worst when using only audio which we hypothesize to be due to the increased difficulty of the task: While speech intonation may be a significant feature for detecting questions in general, discerning between specific questions is easier with access to transcribed keywords.

Including the auxiliary binary classification task (MultiQT-MT) shows no significant improvement over MultiQT. We hypothesize that this may be due to training on a subset of all questions such that there are unlabelled questions in the training data which add noise to the binary task.

Applying tensor fusion instead of concatenating the unimodal representations also does not yield significant improvements to MultiQT contrary to the findings by~\citet{zadeh-etal-2017-tensor}. 
Since tensor-fusion subsumes the concatenated unimodal representations by definition and appends all element-wise products, we must conclude that the multimodal interactions represented by the element-wise products either already exist in the unimodal representations, by correlation, are easily learnable from them or are too difficult to learn for MultiQT.
We believe that the interactions are most likely to be easily learnable from the unimodal representations.

Comparing any MultiQT variant with {\sc instance} and {\sc timestep} F1 clearly shows that {\sc instance} is more forgiving, with models generally achieving higher values in this metric. The difference in performance between different combinations of the modalities is generally higher when measured per {\sc instance} as compared to per {\sc timestep}.

The RF and FNN baseline models clearly underperform compared to MultiQT. It should be noted that both RF and FNN achieve F1-scores of around 85 when evaluated per input utterance, corresponding to the input they receive during training. On this metric, FNN also outperforms RF. However, both models suffer significantly from the discrepancy between the training and streaming settings as measured per the {\sc instance} and {\sc timestep} metrics; this effect is largest for the FNN model.

\paragraph{Real-time tracking.} One important use case of MultiQT is real-time labelling of streamed audio sequences and associated transcripts. For this reason, MultiQT must be able to process a piece of audio in a shorter time than that spanned by the audio itself. For instance, given a \SI{1}{s} chunk of audio, MultiQT must process this in less than \SI{1}{s} in order to maintain a constant latency from the time that the audio is ready to be processed to when it has been processed. To assess the real-time capability of MultiQT, we test it on an average emergency call using an NVIDIA GTX 1080 Ti GPU card. In our data, the average duration of an emergency call is \SI{166}{s}.

To simulate real-time streaming, we first process the call in 166 distinct one-second chunks using 166 sequential forward passes. This benchmark includes all overhead, such as the PCIe transfer of data to and from the GPU for each of the forward passes. The choice of \SI{1}{s} chunk duration matches our production setting but is otherwise arbitrary with smaller chunks giving lower latency and larger chunks giving less computational overhead. In this streaming setting, the \SI{166}{s} of audio are processed in \SI{1.03}{s} yielding a real-time factor of approximately 161 with a processing time of \SI{6.2}{ms} per \SI{1}{s} of audio. This satisfies the real-time constraint by a comfortable margin, theoretically leaving room for up to 161 parallel audio streams to be processed on the same GPU before the real-time constraint is violated.

When a single model serves multiple ongoing calls in parallel, we can batch the incoming audio chunks. Batching further increases the real-time factor and enables a larger number of ongoing calls to be processed in parallel on a single GPU. This efficiency gain comes at the cost of additional, but still constant, latency since we must wait for a batch of chunks to form. For any call, the expected additional latency is half the chunk duration. We perform the same experiment as above but with different batch sizes. We maintain super real-time processing for batches of up 256 one-second chunks, almost doubling the number of calls that can be handled by a single model.

In the offline setting, for instance for on-demand processing of historical recordings, an entire call can be processed in one forward pass. Here, MultiQT can process a single average call of \SI{166}{s} in \SI{10.9}{ms} yielding an offline real-time factor of 15,000. Although batched processing in this setting requires padding, batches can be constructed with calls of similar length to reduce the relative amount of padding and achieve higher efficiency yet.

\section{Discussion}

\paragraph{Label confusion.} We analyze the label confusion of the basic MultiQT model using both modalities on the {\sc timestep} metric. Less than 1\% of all incorrect timestamps correspond to question-to-question confusions while the two primary sources of confusion are incorrect labelings of 1) ``None'' class for a question and 2) of a question with the ``None'' class.
The single highest confusion is between the ``None'' class and ``Q4'' which is the least frequent question. Here the model has a tendency to both over-predict and miss: ca 40\% of predicted ``Q4'' are labeled as ``None'' and 40\% of ``Q4'' are predicted as ``None''. In summary, when our model makes an error, it will most likely 1) falsely predict a non-question to be a question or 2) falsely predict a question to be a non-question; once it discovers a question, it is much less likely to assign it the wrong label.

\paragraph{Model disagreement.} We examined the inter-model agreement between MultiQT trained on the different modes. The highest agreement of $\sim$90\% is achieved between the unimodal text and the multimodal models whereas the lowest agreement was generally between the unimodal audio and any other model at $\sim$80\%. The lower agreement with the unimodal audio model can be attributed to the generally slightly lower performance of this model compared to the other models as per \autoref{tab:results}.

\begin{figure}[t]
    \centering
    \includegraphics[width=\columnwidth]{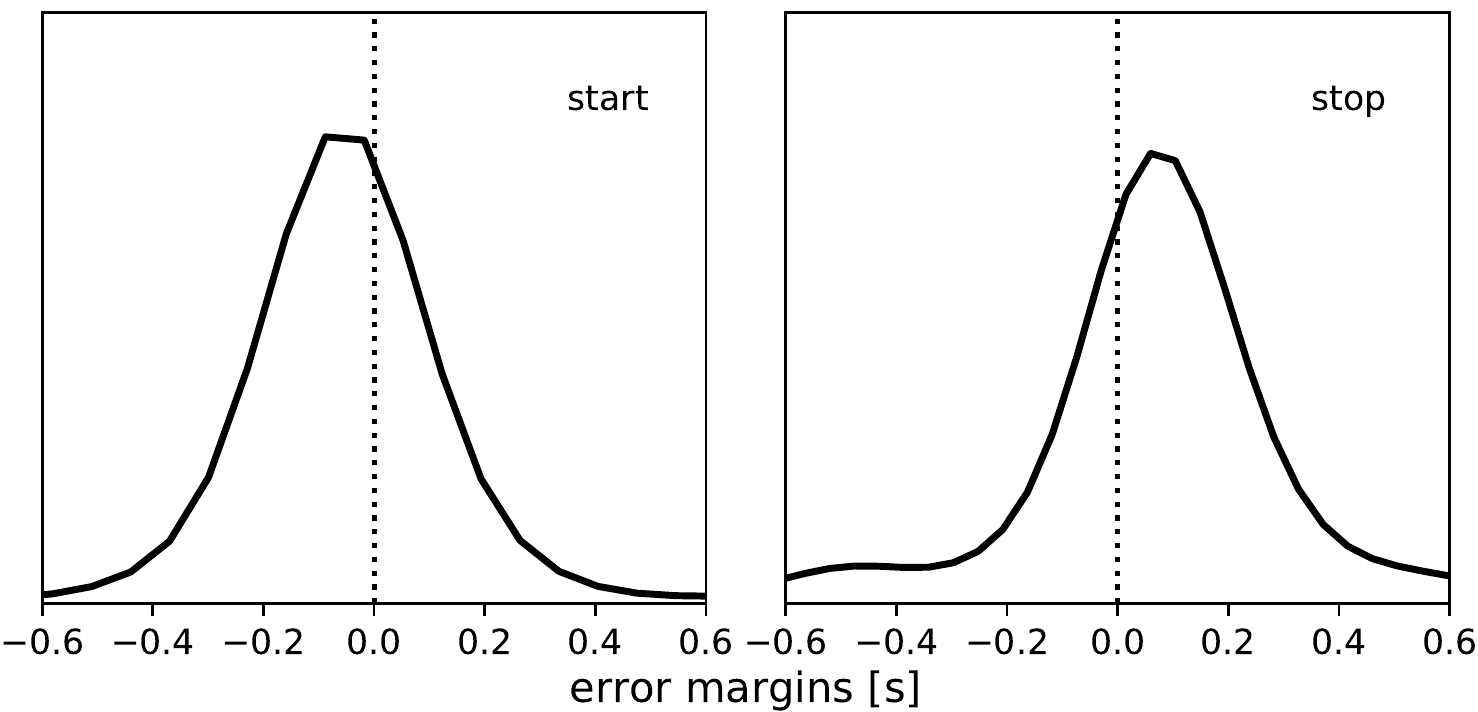}
    \caption{Error margin distributions for start and stop timestamps of question sequences. The dotted lines depict the ground truth start and stop timestamps.}
    \label{fig:errormargins}
\end{figure}

\paragraph{Question margins.} In \autoref{fig:errormargins}, we visualize the distribution of the errors made by the model per {\sc timestep}. For each question regarded as matching according to the {\sc instance} metric we compute the number of seconds by which the model mismatched the label sequence on the left and right side of the label sequence, respectively. We see that the model errors are normally distributed around a center value that is shifted towards the outside of the question by slightly less than \SI{100}{ms}. The practical consequence is that the model tends to make predictions on the \textit{safe side} by extending question segments slightly into the outside of the question.

\begin{table*}
    \centering
    \resizebox{0.9\textwidth}{!}{
        \begin{tabular}{lccrrrrrr}
        \toprule
        & \multicolumn{2}{c}{Permuted} & \multicolumn{3}{c}{{\sc instance}} & \multicolumn{3}{c}{{\sc timestep}}\\ 
        \cmidrule(l{2pt}r{2pt}){2-3}
        \cmidrule(l{2pt}r{2pt}){4-6}
        \cmidrule(l{2pt}r{2pt}){7-9}
        Modality & Training & Test & \multicolumn{1}{c}{P} & \multicolumn{1}{c}{R} & \multicolumn{1}{c}{F1} & \multicolumn{1}{c}{P} & \multicolumn{1}{c}{R} & \multicolumn{1}{c}{F1}\\
        \cmidrule(l{2pt}r{2pt}){1-1}
        \cmidrule(l{2pt}r{2pt}){2-3}
        \cmidrule(l{2pt}r{2pt}){4-6}
        \cmidrule(l{2pt}r{2pt}){7-9}
        A+T & Yes & T & 82.2$\pm$4.9 & 60.1$\pm$5.6 & 68.6$\pm$5.7 & 79.0$\pm$4.7 & 58.4$\pm$3.7 & 64.7$\pm$3.5 \\ 
        A+T & Yes & A & 82.6$\pm$3.2 & 75.9$\pm$2.9 & 78.7$\pm$1.6 & 78.3$\pm$2.4 & 68.3$\pm$2.7 & 72.3$\pm$1.1 \\ 
        A+T & Yes & - & 86.3$\pm$1.6 & 83.8$\pm$2.8 & 84.8$\pm$2.0 & 80.4$\pm$1.0 & 74.1$\pm$2.2 & 76.9$\pm$1.3 \\ 
        \cmidrule(l{2pt}r{2pt}){1-1}
        \cmidrule(l{2pt}r{2pt}){2-3}
        \cmidrule(l{2pt}r{2pt}){4-6}
        \cmidrule(l{2pt}r{2pt}){7-9}
        A+T & No & T & 0.0$\pm$0.0 & 0.0$\pm$0.0 & 0.0$\pm$0.0 & 16.2$\pm$0.0 & 16.7$\pm$0.0 & 16.4$\pm$0.0 \\ 
        A+T & No & A & 89.5$\pm$3.1 & 69.2$\pm$4.4 & 77.0$\pm$2.5 & 84.3$\pm$2.6 & 63.7$\pm$3.5 & 71.0$\pm$2.0 \\ 
        A+T & No & - & 83.6$\pm$2.2 & 83.3$\pm$2.5 & 83.3$\pm$1.6 & 75.7$\pm$2.2 & 73.8$\pm$2.3 & 74.5$\pm$1.3 \\ 
        A & No & - & 87.4$\pm$1.9 & 60.6$\pm$4.0 & 70.3$\pm$3.1 & 79.2$\pm$1.3 & 57.8$\pm$3.3 & 65.0$\pm$2.4 \\ 
        T & No & - & 84.2$\pm$1.6 & 78.5$\pm$2.8 & 81.1$\pm$2.0 & 78.8$\pm$1.2 & 69.4$\pm$2.0 & 73.5$\pm$1.3 \\ 
        \bottomrule
        \end{tabular}
    }
    \caption{Results from the modality ablation on the MultiQT model. We compare multimodal MultiQT trained with the audio (A) and text (T) modalities temporally permuted in turn during training with probability $p_a=0.1$ and $p_s=0.5$ to MultiQT trained without modality permutation, unimodally and multimodally (some results copied from \autoref{tab:results}). We can obtain robustness to loosing a modality while maintaining (or even slightly improving) the multimodal performance.
    All results are based on five-fold cross-validation as in \autoref{tab:results}.
    }
    \label{tab:noise-resilience}
\end{table*}

\begin{figure}[t]
    \centering
    \includegraphics[width=0.8\columnwidth]{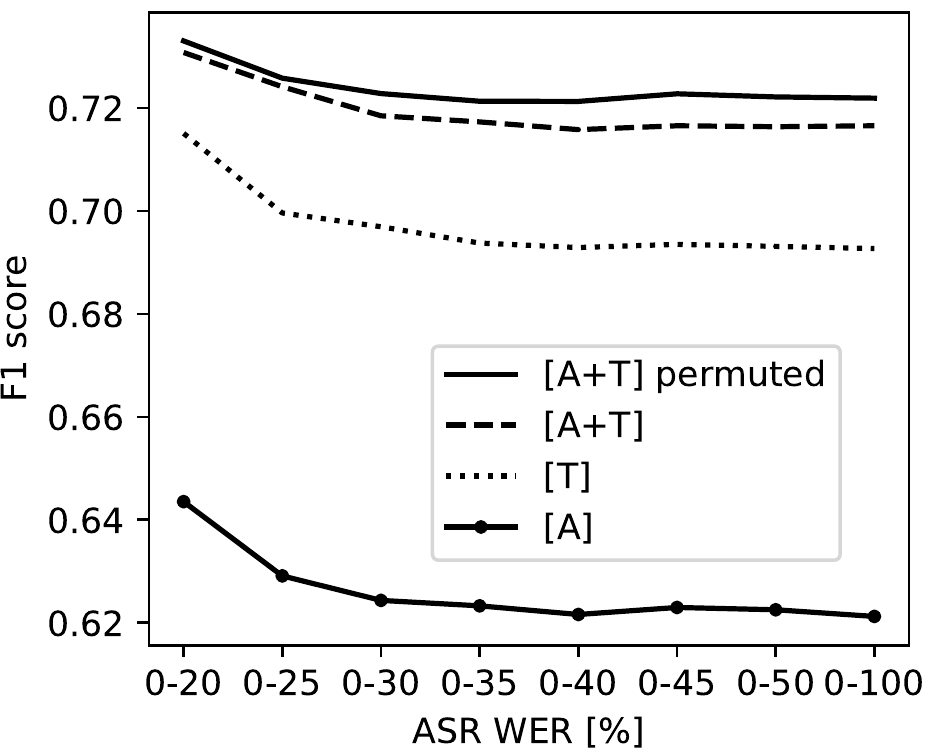}
    \caption{Relation between {\sc timestep} F1 and WER on call-taker utterances without the ``None'' label.}
    \label{fig:wer_plot}
\end{figure}

\paragraph{Modality ablation.} To evaluate the model's robustness to noise in the modalities, we remove all information from one of the modalities in turn and report the results in \autoref{tab:noise-resilience}. 
We remove the information in a modality by randomly permuting the entire temporal axis.
This way we retain the numerical properties of the signal which is not the case when replacing a modality by zeros or noise.
To increase MultiQT's robustness to this modality ablation, we apply it at training so that for each batch example we permute the temporal axis of the audio or text modality with some probability $p_a$ or $p_s$.
We choose $p_a=0.1$ and $p_s=0.5$ since the model more easily develops an over-reliance on the text-modality supposedly due to higher signal-to-noise ratio.
The results are listed in \autoref{tab:noise-resilience} along with results for MultiQT from \autoref{tab:results} for easy reference.
We observe that the basic MultiQT model suffers significantly from permutation of the text modality and less so for audio which suggests that it relies on the audio only for supportive features.
Training MultiQT with the random temporal permutation forces learning of robustness to loosing all information in a modality. We see that the results when removing a modality almost reach the level achieved when training exclusively on that modality while still maintaining the same (or better) performance of the basic MultiQT model.

\paragraph{Relation to ASR.} In \autoref{fig:wer_plot}, we plot the performance of the multimodal model on different subsets of the test split by the maximum WER of the ASR (measured only on the call-taker utterances).
This evaluation compares the micro-averaged model F1-score when increasing the noise on the textual input. We see that regardless of the modality, the performance is the highest for calls with very low WER. We observe that the performance improvement of using both modalities over unimodal text or unimodal audio increases as we include noisy samples. This implies that multi modality increases robustness. Training on permuted inputs additionally improves  the performance on noisy data.

The evaluation of MultiQT in our paper has thus far been only in relation to one particular ASR model with CTC loss~\cite{graves-etal-2006-connectionist}, where our system displays significant gains from multimodal learning. Yet, do these results hold with another ASR system, and in particular, are the multimodal gains still significant if WER decreases and produced text quality increases? For an initial probing of these questions, we replace the fully convolutional ASR with a densely-connected recurrent architecture with convolutional heads. This model is similar to the one in \cite{amodeiDeepSpeechEndtoEnd2015} but also uses dense bottleneck layers. With this model the transcription quality improves by around +4\% in WER, while the F1-scores of MultiQT still strongly favor the multimodal approach, by +6.15 points absolute over text-only. We argue that in a real-world scenario with high WER and limited in-domain training data, the gains warrant learning from joining the text and audio views on the input speech when learning a question tracker. Alternatively, the ASR model itself could be extended into a multitask learning setup to jointly track questions and transcribe speech; we defer that line of work for future research. On a practical note, for this multitask approach, the data must be fully transcribed by human annotators in addition to the question annotatations. This is generally more time consuming and expensive than exclusively annotating questions.

\paragraph{Qualitative analysis.} We analyze the model predictions on a subset of 21 calls to identify the most likely reasons for incorrect labeling. We find that in over half of the analysed cases the incorrect prediction is triggered either by a question-related keyword uttered in a non-question sentence or by a question asked in the background by a caller that was not assigned a label. We also encounter undetected questions that have a very noisy ASR transcript or are asked in an unusual way.

\paragraph{Symptom labeling.} The experiment with our silver-standard symptoms data shows a trend that is similar to question tracking: The dual-modality MultiQT scores an {\sc instance} F1 score of 76.9 for a +1.8 absolute improvement over the best single modality. Text-only is the runner up (-1.8 F1) while audio-only lags behind with a significant -23.6 decrease in F1. At the same time, a simple text-only keyword matching baseline scores at 73.7. We argue that symptom tracking strongly favors text over audio because the distinctive audio features of questions, such as changes in intonation, are not present when communicating symptoms in speech.

\section{Related work}

The broader context of our work is to track the dialogue state in calls to emergency medical services, where conversations are typically formed as sequences of questions and answers that pertain to various medical symptoms. The predominant approach to dialogue state tracking (DST) in speech is to first transcribe the speech by using ASR~\cite{henderson2014second,henderson2015machine,mrksic-etal-2017-neural}. In our specific context, to entirely rely on ASR is prohibitive because of significantly higher WER in comparison to standard datasets. To exemplify, while WER is normally distributed with a mean of 37.6\% in our data, the noisiest DST challenge datasets rarely involve with WER above 30\%~\cite{jagfeld-vu-2017-encoding} while standard ASR benchmarks offer even lower WER~\cite{Park_2019}. None of the standard ASR scenarios thus directly apply to a real-life ASR noise scenario.

From another viewpoint, work in audio recognition mainly involves with detecting simple single-word commands or keyword spotting~\cite{de2018neural}, recognizing acoustic events such as environmental or urban sounds~\cite{salamon2014dataset,piczak2015environmental,xu2016fully} or music patterns, or document-level classification of entire audio sequences~\cite{liu2017topic}. \citet{mcmahan2018listening} provide a more extensive overview. While approaches in this line of work relate to ours, e.g. in the use of convolutional networks over audio~\cite{sainath2015convolutional,salamon2017deep}, our challenge features questions as linguistic units of significantly greater complexity.

Finally, research into multimodal or multi-view deep learning~\cite{ngiam2011multimodal,li2018survey} offers insights to effectively combine multiple data modalities or views on the same learning problem. However, most work does not directly apply to our problem: i) the audio-text modality is significantly under-represented, ii) the models are typically not required to work online, and iii) most tasks are cast as document-level classification and not sequence labeling~\cite{ws-2018-grand}.

\section{Conclusions}
We proposed a novel approach to speech sequence labeling by learning a multimodal representation from the temporal binding of the audio signal and its automatic transcription. This way we learn a model to identify questions in real time with a high accuracy while trained on a small annotated dataset. We show the multimodal representation to be more accurate and more robust to noise than the unimodal approaches. Our findings generalize to a medical symptoms labeling task, suggesting that our model is applicable as a general-purpose speech tagger wherever the speech modality is coupled in real time to ASR output.

\paragraph{Acknowledgements.} The authors are grateful to the anonymous reviewers and area chairs for the incisive and thoughtful treatment of our work.

\bibliography{acl2020}
\bibliographystyle{acl_natbib}

\end{document}